\pdfoutput=1

\documentclass[11pt]{article}

\usepackage[]{acl} 

\usepackage{times}
\usepackage{latexsym}

\usepackage[T1]{fontenc}

\usepackage[utf8]{inputenc}

\usepackage{microtype}

\usepackage{amsmath}
\usepackage{url}
\usepackage{graphicx}
\usepackage{amsfonts}
\usepackage{CJK}
\usepackage{bm}
\usepackage{mathrsfs}
\usepackage{float}
\usepackage{xcolor,colortbl}
\usepackage{subfigure}
\usepackage{booktabs,multirow}
\usepackage{bigstrut,bigdelim}
\usepackage{paralist}
\usepackage{bbm}
\usepackage{helvet}
\usepackage{courier}
\usepackage{makecell}
\usepackage{nicefrac}
\usepackage{epsfig}
\usepackage{diagbox}
\usepackage{framed}
\usepackage{empheq}
\usepackage{array}
\usepackage{enumitem}
\usepackage{mdwlist}
\usepackage{xspace,mfirstuc,tabulary}
\usepackage{tcolorbox}
\usepackage{algorithm}
\usepackage{placeins}
\usepackage{algpseudocode}
\DeclareMathOperator*{\argmax}{argmax}
\DeclareMathOperator*{\softmax}{softmax}

\DeclareMathOperator*{\TransformerEncoder}{TransformerEncoder}

\title{\textsc{HeterMPC}: A Heterogeneous Graph Neural Network for \\ Response Generation in Multi-Party Conversations}

\author{
Jia-Chen Gu$^1$\thanks{\hspace{1.5mm}Work done during the internship at Microsoft.} \ \thanks{\hspace{1.5mm}Equal contribution.}, 
Chao-Hong Tan$^1$\footnotemark[2], Chongyang Tao$^2$, Zhen-Hua Ling$^1$, \\
{\bf Huang Hu$^2$, Xiubo Geng$^2$, Daxin Jiang$^2$\thanks{\hspace{1.5mm}Corresponding author.}} \\
  $^1$National Engineering Research Center for Speech and Language Information Processing, \\
      University of Science and Technology of China, Hefei, China \\
  $^2$Microsoft, Beijing, China \\
{\tt \{gujc,chtan\}@mail.ustc.edu.cn}, {\tt zhling@ustc.edu.cn}, \\ {\tt \{chotao,huahu,xigeng,djiang\}@microsoft.com}
}

\begin{document}
\maketitle
\begin{abstract}
  Recently, various response generation models for two-party conversations have achieved impressive improvements, but less effort has been paid to multi-party conversations (MPCs) which are more practical and complicated.  
  Compared with a two-party conversation where a dialogue context is a sequence of utterances, building a response generation model for MPCs is more challenging, since there exist complicated context structures and the generated responses heavily rely on both interlocutors (i.e., speaker and addressee) and history utterances. 
  To address these challenges, we present HeterMPC, a heterogeneous graph-based neural network for response generation in MPCs which models the semantics of utterances and interlocutors simultaneously with two types of nodes in a graph.
  Besides, we also design six types of meta relations with node-edge-type-dependent parameters to characterize the heterogeneous interactions within the graph.
  Through multi-hop updating, HeterMPC can adequately utilize the structural knowledge of conversations for response generation. 
  Experimental results on the Ubuntu Internet Relay Chat (IRC) channel benchmark show that HeterMPC outperforms various baseline models for response generation in MPCs.
\end{abstract}

\section{Introduction}

\begin{figure}[t]
    \centering
    \includegraphics[width=4.5cm]{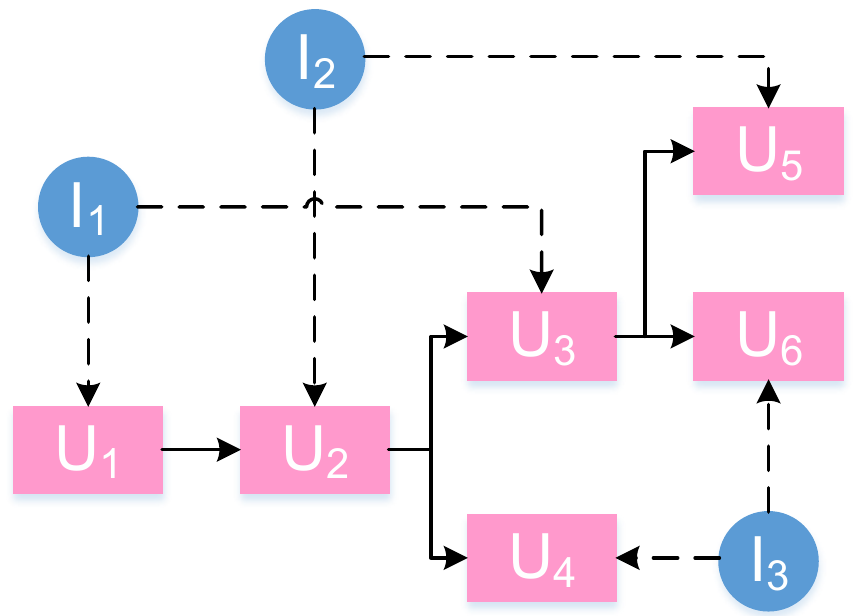}
    \caption{Illustration of a graphical information flow in an MPC.
    Pink rectangles denote utterances and blue circles denote interlocutors. 
    Each solid line represents the ``\emph{replied-by}" relationship between two utterances. Each dashed line indicates the speaker of an utterance. }
    \label{fig-mpc-example}
  \end{figure}

  Enabling dialogue systems to converse naturally with humans is a challenging yet intriguing problem of artificial intelligence and has attracted increasing attention due to its promising potentials and alluring commercial values \cite{DBLP:conf/ccwc/KepuskaB18,DBLP:conf/ucami/Berdasco0D0G19,DBLP:journals/coling/ZhouGLS20}. 
  A large number of researchers have focused on building dialogue generation models with various neural networks.  
  At first, researchers mostly focused on dialogues between two participants \cite{DBLP:conf/acl/ShangLL15,DBLP:conf/aaai/SerbanSBCP16,DBLP:conf/eacl/Rojas-BarahonaG17,DBLP:conf/aaai/YoungCCZBH18}. 
  Recently, researchers have paid more attention to a more practical and challenging scenario involving more than two participants, which is well known as multi-party conversations (MPCs) \cite{DBLP:conf/emnlp/OuchiT16,DBLP:conf/aaai/ZhangLPR18,DBLP:conf/emnlp/LeHSYBZY19,DBLP:conf/ijcai/HuCL0MY19,DBLP:conf/emnlp/WangHJ20,DBLP:conf/acl/GuTLXGJ20}. 
  Utterances in a two-party conversation are posted one by one between two interlocutors, constituting a \emph{sequential} information flow.
  Different from that, utterances in an MPC can be spoken by anyone and address anyone else in this conversation, which constitutes a \emph{graphical} information flow as shown in Figure~\ref{fig-mpc-example}. 
  
  Although sequence-to-sequence (Seq2Seq) models \cite{DBLP:conf/nips/SutskeverVL14,DBLP:conf/aaai/SerbanSBCP16} are effective at modeling sequential dialogues, they fall short of modeling graph-structured ones. 
  To overcome this drawback, \citet{DBLP:conf/ijcai/HuCL0MY19} first proposed a graph-structured network (GSN) to encode utterances based on the graph topology rather than the sequence of their appearances. 
  The graph established in GSN was homogeneous, where nodes represented only utterances. 
  However, interlocutors are also important components of MPCs. 
  There exist complicated interactions between interlocutors, and between an utterance and an interlocutor. 
  Furthermore, when passing messages over a graph, a bidirectional information flow algorithm was designed for GSN. 
  Since both the forward and backward information flows employed the same model structure and parameters, this algorithm cannot distinguish the ``\emph{reply}" or ``\emph{replied-by}" relations between two connected utterance nodes. 
  Also, information flows along both directions are independently propagated, so that a graph node cannot be jointly updated at a single propagation step. 
  
  On account of above issues, we propose a heterogeneous graph-based neural network for response generation in MPCs, named HeterMPC. 
  First, a heterogeneous graph is designed which employs two types of nodes to represent utterances and interlocutors respectively. 
  Different from previous methods that built a homogeneous graph modeling only utterances, utterances and interlocutors are modeled simultaneously in HeterMPC, so that the complicated interactions between interlocutors, between utterances, and between an interlocutor and an utterance can be explicitly described. 
  In order to characterize the heterogeneous attention over each (\emph{source, edge, target}) triple, model parameters dependent on both types of nodes and edges are introduced when calculating  attention weights and passing  messages. 
  Specifically, we introduce six types of meta relations for modeling different edges including ``\emph{reply}'' and ``\emph{replied-by}'' between two utterances, ``\emph{speak}'' and ``\emph{spoken-by}'' between an utterance and a speaker, and ``\emph{address}'' and ``\emph{addressed-by}'' between an utterance and an addressee. 
  With these node-edge-type-dependent structures and parameters, 
  HeterMPC can better utilize the structural knowledge of conversations for node representation and response generation than conventional homogeneous graphs. 
  Finally, Transformer is employed as the backbone of HeterMPC and its model parameters can be initialized with PLMs to take advantage of the recent breakthrough on pre-training.
  
  We evaluate HeterMPC on the Ubuntu Internet Relay Chat (IRC) channel benchmark released by \citet{DBLP:conf/ijcai/HuCL0MY19}. 
  Experimental results show that HeterMPC outperforms GSN \cite{DBLP:conf/ijcai/HuCL0MY19}, GPT-2 \cite{radford2019language}, BERT \cite{DBLP:conf/naacl/DevlinCLT19} and BART \cite{DBLP:conf/acl/LewisLGGMLSZ20} by significant margins in terms of both automated and human evaluation metrics, achieving a new state-of-the-art performance for response generation in MPCs. 

  In summary, our contributions in this paper are three-fold:
  1) To the best of our knowledge, this paper is the first exploration of using heterogeneous graphs for modeling conversations; 
  2) A Transformer-based heterogeneous graph architecture is introduced for response generation in MPCs, in which two types of nodes, six types of meta relations, and node-edge-type-dependent parameters are employed to characterize the heterogeneous properties of MPCs;
  3) Experimental results show that our proposed model achieves a new state-of-the-art performance of response generation in MPCs on the Ubuntu IRC benchmark.

\section{Related Work}

  \paragraph{Multi-Party Conversation}
    Existing methods on building dialogue systems can be generally categorized into generation-based \cite{DBLP:conf/acl/ShangLL15,DBLP:conf/aaai/SerbanSBCP16,DBLP:conf/eacl/Rojas-BarahonaG17,DBLP:conf/aaai/YoungCCZBH18,DBLP:conf/acl/ZhangSGCBGGLD20} or retrieval-based approaches \cite{DBLP:conf/sigdial/LowePSP15,DBLP:conf/acl/WuWXZL17,DBLP:conf/acl/WuLCZDYZL18,DBLP:conf/wsdm/TaoWXHZY19,DBLP:conf/acl/TaoWXHZY19,DBLP:conf/emnlp/GuLZL19,DBLP:conf/cikm/GuLLLSWZ20}. 
    In this paper, we study the task of response generation in MPCs, where in addition to utterances, interlocutors are also important components who play the roles of speakers or addressees.
    Previous methods have explored retrieval-based approaches for MPCs. 
    For example, \citet{DBLP:conf/emnlp/OuchiT16} proposed the dynamic model which updated speaker embeddings with conversation streams. 
    \citet{DBLP:conf/aaai/ZhangLPR18} proposed speaker interaction RNN which updated speaker embeddings role-sensitively. 
    \citet{DBLP:conf/emnlp/WangHJ20} proposed to track the dynamic topic in a conversation. 
    \citet{DBLP:conf/acl/GuTLXGJ20} proposed jointly learning ``who says what to whom" in a unified framework by designing self-supervised tasks during pre-training. 
    On the other hand, \citet{DBLP:conf/ijcai/HuCL0MY19} explored generation-based approaches by proposing a graph-structured network, the core of which was an utterance-level graph-structured encoder. 

  \paragraph{Heterogeneous Graph Neural Network}
    Early studies on graph neural networks (GNNs) focused on homogeneous graphs where a whole graph is composed of a single type of nodes. 
    However, graphs in real-world applications usually come with multiple types of nodes, namely heterogeneous information networks (HINs) or heterogeneous graphs~\cite{DBLP:series/synthesis/2012Sun}. 
    Recently, researchers have attempted to extend GNNs to model heterogeneity. 
    For example, \citet{DBLP:conf/kdd/ZhangSHSC19} adopted different RNNs for different types of nodes to integrate multi-modal features. 
    \citet{DBLP:conf/www/WangJSWYCY19} extended graph attention networks by maintaining different weights for different meta-path-defined edges. 
    \citet{DBLP:conf/www/HuDWS20} proposed heterogeneous graph Transformer (HGT) to model heterogeneity by maintaining dedicated representations for different types of nodes and edges. 
    In addition, heterogeneous graphs have also been applied to many NLP tasks, such as multi-hop reading comprehension \cite{DBLP:conf/acl/TuWHTHZ19}, text classification \cite{DBLP:conf/emnlp/HuYSJL19} and document summarization \cite{DBLP:conf/acl/WangLZQH20}. 
    
    Previous studies have verified the superiority of modeling MPCs with homogeneous graphs considering only utterances. 
    We claim that it is indeed necessary to model a complex information flow in MPCs shown in Figure~\ref{fig-mpc-example} with a heterogeneous graph, since a homogeneous one cannot explicitly model the relationships of multiple utterances spoken by or addressing an interlocutor. 
    Nowadays, HINs have been widely used in many NLP tasks. 
    To the best of our knowledge, this paper makes the first attempt to build a heterogeneous graph-based neural network considering utterances and interlocutors simultaneously for response generation in MPCs. 
    In addition, we introduce many task-specific modelings for MPCs such as graph construction and node updating which will be elaborated in the model section.

\section{Problem Formulation} \label{sec:pf}
  The task of response generation in MPCs is to generate an appropriate response $\bar{r}$ given the conversation history, the speaker of a response, and which utterance the response is going to reply to, which can be formulated as:
  \begin{equation} \label{equ: generate words}
  \begin{aligned}
    \bar{r} =& \argmax_r log P(r|\mathbb{G}) \\
            =& \argmax_r \sum_{k=1}^{|r|} log P(r_k|\mathbb{G} r_{<k}).
  \end{aligned}
  \end{equation}
  Here, $\mathbb{G}$ is a heterogeneous graph containing both history conversation and the response to be generated. 
  The speaker and addressee of the response are known and its contents are masked.
  The response tokens are generated in an auto-regressive way. 
  $r_k$ and $r_{<k}$ stand for the $k$-th token and the first $(k-1)$ tokens of response $r$ respectively. $|r|$ is the length of $r$.
  
  We will introduce how to construct the graph and how to model the probability in Eq.~(\ref{equ: generate words}) given the built graph in the next section. 

\section{HeterMPC Model} \label{sec-model}

  \begin{figure}[t]
    \centering
    \includegraphics[width=6cm]{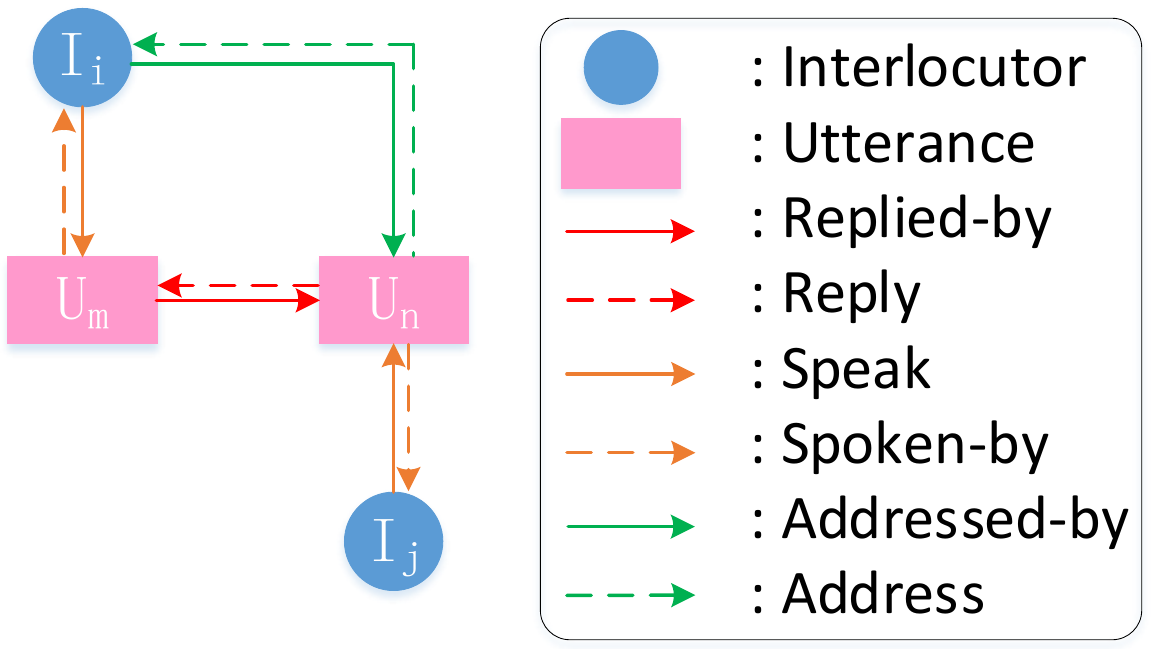}
    \caption{Illustration of the two types of nodes and six types of edges in a heterogeneous conversation graph. 
    This example demonstrates that I$_j$ speaks U$_n$ replying U$_m$ that is spoken-by I$_i$.}
    \vspace{-4mm}
    \label{fig-mpc-edges}
  \end{figure}

  \begin{figure*}[t]
    \centering
    \includegraphics[width=15cm]{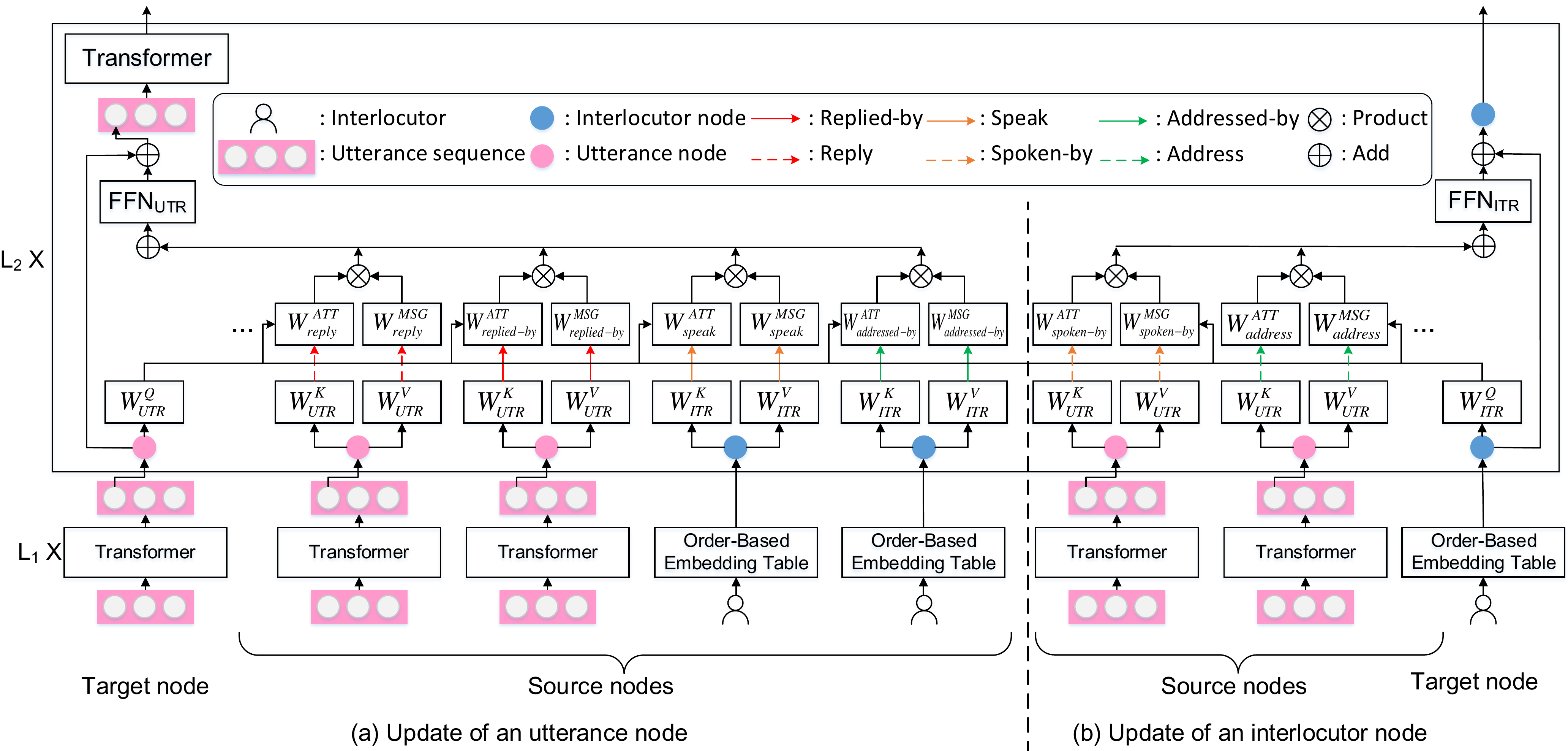}
    \caption{Model architecture of HeterMPC for (a) update of an utterance node and (b) update of an interlocutor node. ``UTR" and ``ITR" are abbreviations of ``utterance" and ``interlocutor" respectively. The set of $W_*^*$ denotes the node-edge-type-dependent parameters.}
    \vspace{-4mm}
    \label{fig-model}
  \end{figure*}

  HeterMPC adopts an encoder-decoder architecture consisting of stacked encoder and decoder layers for graph-to-sequence learning~\cite{DBLP:conf/acl/YaoWW20}. 
  The graph encoder is designed to capture conversation structures and output the representations of all nodes in a graph that are fed to the decoder for response generation. 
  
  \subsection{Graph Construction}
    A heterogeneous graph is constructed to explicitly model the complicated interactions between interlocutors, between utterances, and between an interlocutor and an utterance in an MPC. 
    This graph models utterances and interlocutors by considering them as two types of nodes under a unified framework.    
    Given an MPC instance composed of $M$ utterances and $I$ interlocutors, a heterogeneous graph $\mathbb{G}(\mathbb{V}, \mathbb{E})$ is constructed.
    Specifically, $\mathbb{V}$ is a set of $M+I$ nodes.
    Each node denotes either an utterance or an interlocutor. 
    $\mathbb{E} = \{e_{p,q}\}_{p,q=1}^{M+I}$ is a set of directed edges. 
    Each edge $e_{p,q}$ describes the connection from  node $p$ to node $q$.
  
    Inspired by~\citet{DBLP:journals/pvldb/SunHYYW11,DBLP:conf/kdd/SunNHYYY12}, we introduce six types of meta relations \{\emph{reply}, \emph{replied-by}, \emph{speak}, \emph{spoken-by}, \emph{address}, \emph{addressed-by}\} to describe the directed edge between two graph nodes as illustrated in Figure~\ref{fig-mpc-edges}.
    Specifically, if an utterance represented by node $n$ replies another utterance represented by node $m$, the edge $e_{n,m}=\emph{reply}$ and the reversed edge $e_{m,n}=\emph{replied-by}$. 
    If an utterance represented by node $m$ is spoken by an interlocutor represented by node $i$, $e_{i,m}=\emph{speak}$ and $e_{m,i}=\emph{spoken-by}$.
    If an utterance represented by node $n$ addresses an interlocutor represented by node $i$, $e_{n,i}=\emph{address}$ and $e_{i,n}=\emph{addressed-by}$.
    In other cases, $e_{p,q}=\emph{NULL}$ indicating that there is no connection between these two nodes. 
    Note that it is necessary to distinguish the bidirectional edges between every two nodes that indicate the active and passive tense information respectively.

  \subsection{Node Initialization}
    In HeterMPC, each node is represented as a vector.
    These vectors are first initialized individually without considering graph edges.

    \paragraph{Utterances} 
    When encoding utterances, a \texttt{[CLS]} token is inserted at the start of each utterance, denoting the utterance-level representation. 
    Besides, a \texttt{[SEP]} token is also inserted at the end of each utterance~\cite{DBLP:conf/naacl/DevlinCLT19}.
    Then each utterance is encoded individually by stacked Transformer encoder layers through the self-attention mechanism to derive the contextualized utterance representations.\footnote{In our experiments, BERT or BART was selected to initialize the utterance encoder layers of HeterMPC. Then, the built HeterMPC models were compared with the baseline models directly finetuning BERT or BART, respectively. It is worth noting that the utterance encoder layers of HeterMPC can also be initialized by other types of PLMs, and the comparison across PLMs is not the focus of this paper.}
    The output of a Transformer encoder layer is used as the input of the next layer. 
    Readers can refer to \citet{DBLP:conf/nips/VaswaniSPUJGKP17} for details of Transformer. 
    Formally, the calculation for an utterance at the \emph{l}-th Transformer layer is denoted as:
    \begin{align}
      \boldsymbol{H}_m^{l+1} = \mathop{\TransformerEncoder}( \boldsymbol{H}_m^l ),
    \end{align}
    where $m \in \{1, ..., M\}$, $l \in \{0, ..., L_1 - 1\}$, $L_1$ denotes the number of Transformer layers for initialization, $\boldsymbol{H}_m^l \in \mathbb{R}^{k_m \times d}$, $k_m$ denotes the length of an utterance  and $d$ denotes the dimension of embedding vectors.

    \paragraph{Interlocutors}
    Different from an utterance composed of a sequence of tokens, an interlocutor is directly represented with an embedding vector. 
    Interlocutors in a conversation are indexed according to their speaking order and the embedding vector for each interlocutor is derived by looking up an order-based interlocutor embedding table~\cite{DBLP:conf/cikm/GuLLLSWZ20} that is updated during end-to-end learning. 
    The first interlocutors in all conversation sessions share the same embedding vector in the interlocutor embedding table, so do all the second interlocutors.\footnote{In our experiments, the maximum interlocutor number was set to 10 and an embedding table sized 10*768 was learned during training. We did study initializing the embedding vector of an interlocutor node by averaging the representations of all utterance nodes it speaks, but no further improvement can be achieved.} 
    Thus, this order-based embedding table can be shared across the training, validation and testing sets, and there is no need to estimate an embedding vector for each specific interlocutor in the dataset.

  \subsection{Node Updating}
    As shown in Figure~\ref{fig-model}, the initialized node representations are updated by feeding them into the built graph for absorbing context information~\cite{DBLP:conf/iclr/KipfW17,DBLP:conf/iclr/VelickovicCCRLB18,DBLP:conf/nips/YunJKKK19}.
    We calculate heterogeneous attention weights between connected nodes and pass messages over the graph in a node-edge-type-dependent manner, inspired by introducing parameters to maximize feature distribution differences for modeling heterogeneity~\cite{DBLP:conf/esws/SchlichtkrullKB18,DBLP:conf/www/WangJSWYCY19,DBLP:conf/kdd/ZhangSHSC19,DBLP:conf/www/HuDWS20}.
    After collecting the information from all source nodes to a target node, a node-type-dependent feed-forward network (FFN) followed by a residual connection~\cite{DBLP:conf/cvpr/HeZRS16} is employed to aggregate the information. 
    Then, in order to let each token in an utterance  have access to the information from other utterances, an additional Transformer layer is placed  for utterance nodes specifically. 
    $L_2$ denotes the number of iterations for updating both utterance and interlocutor nodes.

    \subsubsection{Heterogeneous Attention}
      Since the representations of two types of nodes are initialized in different ways, node-type-dependent linear transformations are first applied to node representations before attention calculation so that the two types of nodes share similar feature distributions~\cite{DBLP:conf/www/WangJSWYCY19,DBLP:conf/www/HuDWS20}. 
      Meanwhile, each of the six relation types is assigned a separate linear projection so that the semantic relationship between two connected nodes can be accurately described when calculating attention weights. 
      The forward and backward information flows between them can also be distinguished. 
      
      Formally, let the triple (\emph{s, e, t}) denote an edge \emph{e} connecting a source node \emph{s} to a target node \emph{t}.
      The representations of the source and target nodes at the \emph{l}-th iteration\footnote{For an utterance, the representation for the \texttt{[CLS]} token is extracted as the utterance-level representation.} are denoted as $\boldsymbol{h}^l_s$ and $\boldsymbol{h}^l_t$, serving as a \emph{key} ($K$) vector and a \emph{query} ($Q$) vector of attention calculation respectively.
      Then, the heterogeneous attention weight $w^l(s, e, t)$ before normalization for this triple is calculated as:
      \begin{align}
        \boldsymbol{k}^l(s) &= \boldsymbol{h}^l_s \boldsymbol{W}^K_{\tau(s)} + \boldsymbol{b}^K_{\tau(s)}, \label{equ-att-k} \\
        \boldsymbol{q}^l(t) &= \boldsymbol{h}^l_t \boldsymbol{W}^Q_{\tau(t)} + \boldsymbol{b}^Q_{\tau(t)}, \label{equ-att-q} \\
        w^l(s, e, t) &= \boldsymbol{k}^l(s) \boldsymbol{W}^{ATT}_{e_{s,t}} {\boldsymbol{q}^l(t)}^T  \frac{\mu_{e_{s,t}}}{\sqrt{d}}. \label{equ-att-qk}
      \end{align}
      Here, $\tau(s), \tau(t) \in$ \emph{\{UTR, ITR\}} distinguish utterance (\emph{UTR}) and interlocutor (\emph{ITR}) nodes. 
      Eqs.~(\ref{equ-att-k}) and (\ref{equ-att-q}) are node-type-dependent linear transformations.
      Eq.~(\ref{equ-att-qk}) contains an edge-type-dependent linear projection $\boldsymbol{W}^{ATT}_{e_{s,t}}$ where
      $\mu_{e_{s,t}}$ is an adaptive factor scaling to the attention. 
      All $\boldsymbol{W}^* \in \mathbb{R}^{d \times d}$ and $\boldsymbol{b}^* \in \mathbb{R}^d$ are parameters to be learnt. 

    \subsubsection{Heterogeneous Message Passing}
      When passing the message of a source node that serves as a \emph{value} ($V$) vector to a target node, node-edge-type-dependent parameters are also introduced considering the heterogeneous properties of nodes and edges. 
      Mathematically:
      \begin{align}
        \boldsymbol{v}^l(s) &= \boldsymbol{h}^l_s \boldsymbol{W}^V_{\tau(s)} + \boldsymbol{b}^V_{\tau(s)}, \\
        \boldsymbol{\bar{v}}^l(s) &= \boldsymbol{v}^l(s) \boldsymbol{W}^{MSG}_{e_{s,t}},
      \end{align}
    where $\boldsymbol{\bar{v}}^l(s)$ is the passed message and all $\boldsymbol{W}^* \in \mathbb{R}^{d \times d}$ and $\boldsymbol{b}^* \in \mathbb{R}^d$ are parameters to be learnt.

    \subsubsection{Heterogeneous Aggregation}
      For a target node, the messages passed from all its connected source nodes need to be aggregated.
      A softmax function is applied to normalize the attention weights and then the messages from all source codes are summarized as:
      \begin{align}
        \boldsymbol{\bar{h}}^l_t  = \sum_{s \in S(t)}\mathop{\softmax} (w^l(s, e, t))  \boldsymbol{\bar{v}}^l(s), 
      \end{align}
      where $S(t)$ denotes the set of source nodes for the target node $t$.
      Then the summarized message $\boldsymbol{\bar{h}}^l_t$ is aggregated with the original node representation $\boldsymbol{h}^l_t$ using a node-type-dependent FFN followed by a residual connection~\cite{DBLP:conf/cvpr/HeZRS16} as:
      \begin{align}
        \boldsymbol{h}^{l+1}_t = \emph{FFN}_{\tau(t)}( \boldsymbol{\bar{h}}^l_t ) + \boldsymbol{h}^l_t, \label{equ-ffn}
      \end{align}
      where the output $\boldsymbol{h}^{l+1}_t$ is used as the input of the next iteration of node updating. 
      One iteration can be viewed as a single-step information propagation along edges. 
      When stacking $L_2$ iterations, a node can attend to other nodes up to $L_2$ hops away. 
      
      A specific consideration on utterance nodes is that the tokens except  \texttt{[CLS]} in an utterance have no access to other utterances during the node updating process introduced above.
      To overcome this disadvantage and derive more contextualized utterance representations, an additional Transformer layer~\cite{DBLP:conf/nips/VaswaniSPUJGKP17} is further placed for utterance nodes as shown in Figure~\ref{fig-model}. 
      In detail, at the $l$-th iteration, the representations of an utterance node before and after node updating, i.e., $\boldsymbol{h}^l_t$ and $\boldsymbol{h}^{l+1}_t$, are concatenated
     and then compressed by a linear transformation as: 
      \begin{align}
       \boldsymbol{\hat{h}}^{l+1}_t = [\boldsymbol{h}^l_t;\boldsymbol{h}^{l+1}_t] \boldsymbol{W}_{com} + \boldsymbol{b}_{com},
      \end{align}
      where $\boldsymbol{W}_{com} \in \mathbb{R}^{2d \times d}$ and $\boldsymbol{b}_{com} \in \mathbb{R}^d$ are parameters. 
      Then, $\boldsymbol{\hat{h}}^{l+1}_t$ replaces the representation of \texttt{[CLS]} (i.e., $\boldsymbol{h}^l_t$) in the sequence representations of the whole utterance.
      Finally, the updated sequence representations are fed into the additional Transformer layer for another round of intra-utterance self-attention, so that 
      the context information learnt by the \texttt{[CLS]} representation can be shared with other tokens in the utterance.

  \subsection{Decoder}
    
    \begin{figure}[t]
      \centering
      \includegraphics[width=5cm]{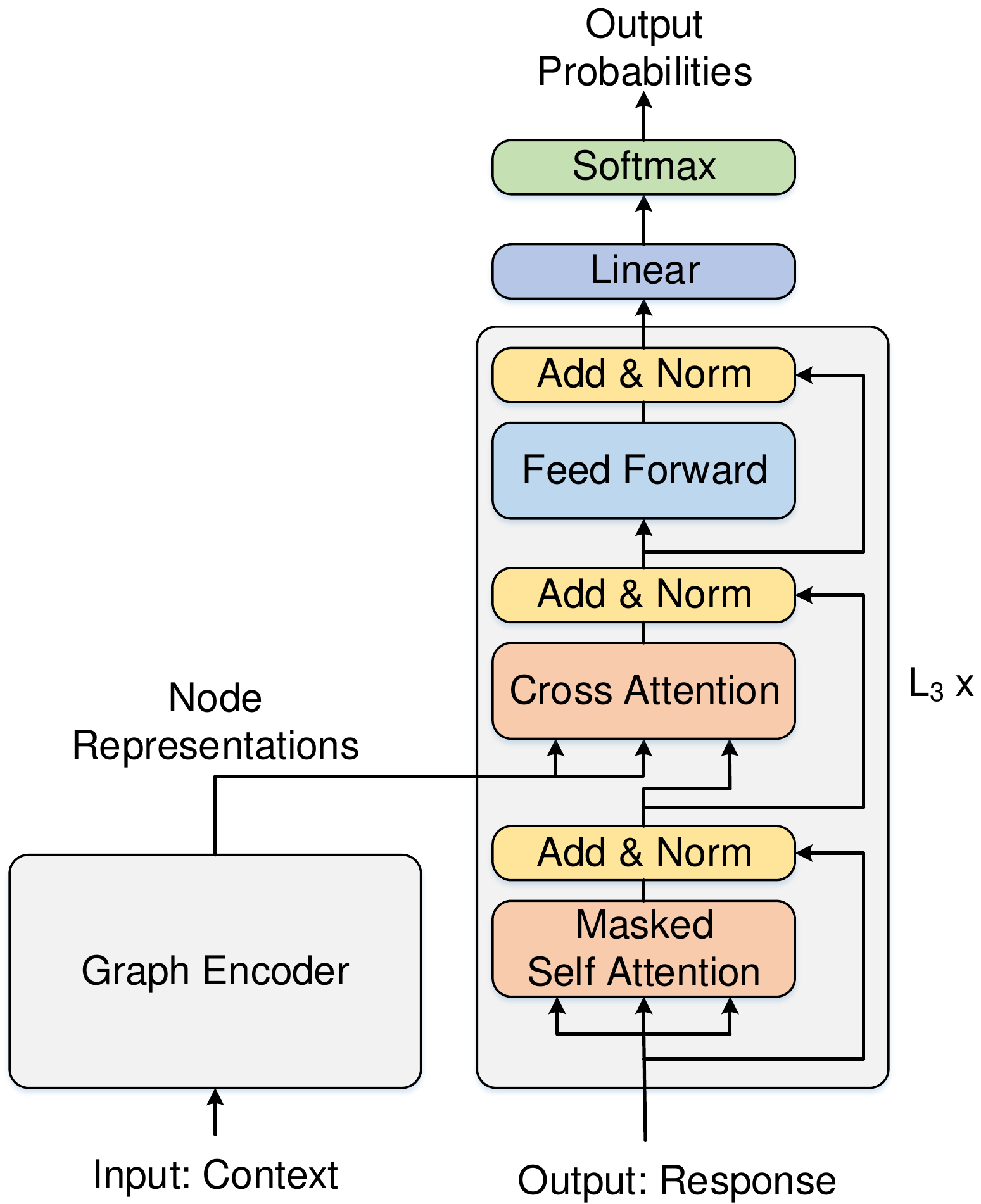}
      \caption{The decoder architecture of HeterMPC.}
      \vspace{-4mm}
      \label{fig-decoder}
    \end{figure}
    
    The decoder is composed of a stack of identical layers as shown in Figure~\ref{fig-decoder}. 
    We follow the standard implementation of Transformer decoder to generate responses. 
    In each decoder layer, a masked self-attention operation is first performed where each token cannot attend to future tokens to avoid information leakage. 
    Furthermore, a cross-attention operation over the node representations of the graph encoder output is performed to incorporate graph information for decoding.
    It is notable that a residual connection along with layer normalization is followed by each attention operation.

\section{Experiments}
  
    \begin{table*}[t]
      \centering
      \setlength{\tabcolsep}{5.0pt}
      \resizebox{0.9\linewidth}{!}{
      \begin{tabular}{lcccccc}
      \toprule
        \backslashbox{Models}{Metrics}                     &  BLEU-1 &  BLEU-2 &  BLEU-3 &  BLEU-4 &  METEOR &  ROUGE$_L$  \\
      \midrule
        Seq2Seq (LSTM) \cite{DBLP:conf/nips/SutskeverVL14} &  7.71   &  2.46   &  1.12   &  0.64   &  3.33   &  8.68   \\
        Transformer \cite{DBLP:conf/nips/VaswaniSPUJGKP17} &  7.89   &  2.75   &  1.34   &  0.74   &  3.81   &  9.20  \\
        GSN \cite{DBLP:conf/ijcai/HuCL0MY19}               &  10.23  &  3.57   &  1.70   &  0.97   &  4.10   &  9.91  \\
        GPT-2 \cite{radford2019language}                   &  10.37  &  3.60   &  1.66   &  0.93   &  4.01 	 &  9.53  \\
      \midrule
        BERT \cite{DBLP:conf/naacl/DevlinCLT19}            &  10.90  &  3.85   &  1.69   &  0.89   &  4.18   &  9.80  \\
        HeterMPC$_{\emph{BERT}}$                           &  \textbf{12.61}  &  \textbf{4.55}   &  \textbf{2.25}   &  \textbf{1.41}   &  \textbf{4.79}   &  \textbf{11.20}  \\
        HeterMPC$_{\emph{BERT}}$ w/o. node types           &  11.76  &  4.09   &  1.87   &  1.12   &  4.50   &  10.73  \\
        HeterMPC$_{\emph{BERT}}$ w/o. edge types           &  12.02  &  4.27   &  2.10   &  1.30   &  4.74   &  10.92  \\
        HeterMPC$_{\emph{BERT}}$ w/o. node and edge types  &  11.60  &  3.98   &  1.90   &  1.18   &  4.20   &  10.63  \\
        HeterMPC$_{\emph{BERT}}$ w/o. interlocutor nodes   &  11.80  &  3.96   &  1.75   &  1.00   &  4.31   &  10.53  \\
      \midrule
        BART \cite{DBLP:conf/acl/LewisLGGMLSZ20}           &  11.25  &  4.02   &  1.78   &  0.95   &  4.46   &  9.90  \\
        HeterMPC$_{\emph{BART}}$                           &  \textbf{12.26}  &  \textbf{4.80}   &  \textbf{2.42}   &  \textbf{1.49}   &  \textbf{4.94}   &  \textbf{11.20}  \\
        HeterMPC$_{\emph{BART}}$ w/o. node types           &  11.22  &  4.06   &  1.87   &  1.04   &  4.57   &  10.63  \\
        HeterMPC$_{\emph{BART}}$ w/o. edge types           &  11.52  &  4.27   &  2.05   &  1.24   &  4.78   &  10.90  \\
        HeterMPC$_{\emph{BART}}$ w/o. node and edge types  &  10.90  &  3.90   &  1.79   &  1.01   &  4.52   &  10.79  \\
        HeterMPC$_{\emph{BART}}$ w/o. interlocutor nodes   &  11.68  &  4.24   &  1.91   &  1.03   &  4.79   &  10.65  \\
      \bottomrule
      \end{tabular}
      }
      \caption{Performance of HeterMPC and ablations on the test set in terms of automated evaluation. Numbers in bold denote that the improvement over the best performing baseline is statistically significant (t-test with \emph{p}-value $<$ 0.05).}
      \label{tab-result-1}
    \end{table*}

    
    \begin{table}[t]
      \centering
      \setlength{\tabcolsep}{5.0pt}
      \resizebox{0.9\linewidth}{!}{
      \begin{tabular}{lcc}
      \toprule
        \backslashbox{Models}{Metrics}           &   Score  &   Kappa \\
      \midrule
        Human                                    &   2.81   &   0.55 \\ 
      \midrule
        GSN \cite{DBLP:conf/ijcai/HuCL0MY19}     &   2.00   &   0.50 \\ 
        BERT \cite{DBLP:conf/naacl/DevlinCLT19}  &   2.19   &   0.42 \\ 
        BART \cite{DBLP:conf/acl/LewisLGGMLSZ20} &   2.24   &   0.44 \\ 
        HeterMPC$_{\emph{BERT}}$                 &   2.39   &   0.50 \\ 
        HeterMPC$_{\emph{BART}}$                 &   2.41   &   0.45 \\ 
      \bottomrule
      \end{tabular}
      }
      \caption{Human evaluation results of HeterMPC and some selected systems on a randomly sampled test set.}
      \label{tab:Human evaluation}
    \end{table}

  \subsection{Datasets}
    
    We evaluated our proposed method on the Ubuntu IRC benchmark used in \citet{DBLP:conf/ijcai/HuCL0MY19}. 
    The data processing script provided by \citet{DBLP:conf/ijcai/HuCL0MY19} was employed to derive the dataset.\footnote{We contacted the authors of \citet{DBLP:conf/ijcai/HuCL0MY19} to obtain the data processing script. As they claimed, it was an updated version which was a little different from that used in their paper. Thus, we re-implemented all baselines on this updated dataset to ensure fair comparison.}
    In this dataset, both speaker and addressee labels were included for each utterance in a session. 
    When testing, the speaker and addressee information was both given for response generation, i.e., the system knew who would speak next and which utterance should be responded to following the graph structure.
    It contained 311,725/5,000/5,000 dialogues in the training/validation/testing sets respectively. 

  \subsection{Baseline Models}

    We compared our proposed methods with as many MPC models as possible.
    Considering that there are only a few research papers in this field, several recent advanced models were also adapted to provide sufficient comparisons.
    Finally, we compared with the following baseline models:
    \textbf{(1) RNN-based Seq2Seq} \cite{DBLP:conf/nips/SutskeverVL14} took all utterances except the target utterance to generate as input, which were sorted according to their posting time and concatenated. Thus, structured conversations were converted into sequential ones. Seq2Seq modeling with attention was performed as that in \citet{DBLP:conf/nips/SutskeverVL14,DBLP:journals/corr/BahdanauCB14} on the concatenated utterances.
    \textbf{(2) Transformer} \cite{DBLP:conf/nips/VaswaniSPUJGKP17} took the same input utterances as those used for the Seq2Seq model.
    \textbf{(3) GPT-2} \cite{radford2019language} was a uni-directional pre-trained language model. Following its original concatenation operation, all context utterances and the response were concatenated with a special \texttt{[SEP]} token as input for encoding. 
    \textbf{(4) BERT} \cite{DBLP:conf/naacl/DevlinCLT19} concatenated all context utterances and the response similarly as those for GPT-2. To adapt BERT for response generation, a special masking mechanism was designed to avoid response information leakage during encoding. Concretely, each token in the context utterances attended to all tokens in the context utterances, while each token in the response cannot attend to future tokens in the utterance. 
    \textbf{(5) GSN} \cite{DBLP:conf/ijcai/HuCL0MY19} achieved the state-of-the-art performance on MPCs. The core of GSN was an utterance-level graph-structured encoder.
    \textbf{(6) BART} \cite{DBLP:conf/acl/LewisLGGMLSZ20} was a denoising autoencoder using a standard Tranformer-based architecture, trained by corrupting text with an arbitrary noising function and learning to reconstruct the original text. In our experiments, a concatenated context started with <s> and separated with </s> were fed into the encoder, and a response were fed into the decoder.

  \subsection{Evaluation Metrics}
    To ensure all experimental results were comparable, we used the same automated and human evaluation metrics as those used in previous work \cite{DBLP:conf/ijcai/HuCL0MY19}. 
    \citet{DBLP:conf/ijcai/HuCL0MY19} used the evaluation package released by \citet{DBLP:journals/corr/ChenFLVGDZ15} including BLEU-1 to BLEU-4, METEOR and ROUGE$_L$, which was also used in this paper.\footnote{https://github.com/tylin/coco-caption} 
    Human evaluation was conducted to measure the quality of the generated responses in terms of three independent aspects: 1) relevance, 2) fluency and 3) informativeness. 
    Each judge was asked to give three binary scores for a response, which were further summed up to derive the final score ranging from 0 to 3.

  \subsection{Training Details}
    Model parameters were initialized with pre-trained weights of \emph{bert-base-uncased} released by \citet{DBLP:conf/emnlp/WolfDSCDMCRLFDS20}. 
    The AdamW method~\cite{DBLP:conf/iclr/LoshchilovH19} was employed for optimization. 
    The learning rate was initialized as $6.25e\text{-}5$ and was decayed linearly down to $0$. 
    The max gradient norm was clipped down to $1.0$.
    The batch size was set to $16$ with $8$ gradient accumulation steps.
    The maximum utterance length was set to $50$. 
    The number of layers for initializing utterance representations $L_1$ was set to 9, and the number of layers for heterogeneous graph iteration $L_2$ was set to 3. 
    $L_1$ and $L_2$ were validated on the validation set.
    The number of decoder layers $L_3$ was set to 6, achieving the best performance out of \{2, 4, 6, 8\} on the validation set.
    The strategy of greedy search was performed for decoding. 
    The maximum length of responses for generation was also set to $50$. 
    All experiments were run on a single GeForce RTX 2080 Ti GPU. 
    The maximum number of epochs was set to 15, taking about 40 hours. 
    The validation set was used to select the best model for testing. 
    All code was implemented in the PyTorch framework\footnote{https://pytorch.org/} and are published to help replicate our results.~\footnote{https://github.com/lxchtan/HeterMPC}

  \subsection{Evaluation Results}
    In our experiments, BERT and BART were selected to initialize HeterMPC.
    HeterMPC$_{\emph{BERT}}$ denoted that the utterance encoder was initialized with BERT and the decoder was randomly initialized.
    HeterMPC$_{\emph{BART}}$ denoted the encoder and decoder were initialized by those of BART respectively.

    \paragraph{Automated Evaluation}
    Table~\ref{tab-result-1} presents the evaluation results of HeterMPC$_{\emph{BERT}}$, HeterMPC$_{\emph{BART}}$ and previous methods on the test set. 
    Each model ran four times with identical architectures and different random initializations, and the best out of them was reported. 
    We ran the code released by \citet{DBLP:conf/ijcai/HuCL0MY19} to reproduce the results of GSN for a fair comparison.\footnote{https://github.com/morning-dews/GSN-Dialogues}
    The results show that both HeterMPC$_{\emph{BERT}}$ and HeterMPC$_{\emph{BART}}$ outperformed all baselines in terms of all metrics. 
    HeterMPC$_{\emph{BERT}}$ outperformed GSN by 2.38\% BLEU-1 and 0.44\% BLEU-4, and outperformed GPT-2 by 2.24\% BLEU-1 and 0.48\% BLEU-4.
    HeterMPC$_{\emph{BART}}$ outperformed GSN by 2.03\% BLEU-1 and 0.52\% BLEU-4, and outperformed GPT-2 by 1.89\% BLEU-1 and 0.56\% BLEU-4.
    Furthermore, HeterMPC$_{\emph{BERT}}$ outperformed BERT by 1.71\% BLEU-1 and 0.52\% BLEU-4, and HeterMPC$_{\emph{BART}}$ outperformed BART by 1.01\% BLEU-1 and 0.54\% BLEU-4, illustrating the importance of modeling MPC structures. 
    
    To further verify the effectiveness of our proposed methods, ablation tests were conducted as shown in Table~\ref{tab-result-1}. 
    First, all nodes or edges were considered equivalently by employing the same linear transformations in Eqs.~(\ref{equ-att-k}) to (\ref{equ-ffn}) for all node or edge types without distinguishing them. 
    The drop in performance illustrates the effectiveness of the node-edge-type-dependent parameters. 
    On the other hand, interlocutor nodes were removed out of a graph and only the meta relations of \emph{reply} and \emph{replied-by} were left. 
    The drop in performance illustrates the importance of modeling interactions between utterances and interlocutors, and the effectiveness of the heterogeneous architecture.
    
    \paragraph{Human Evaluation}
    Table~\ref{tab:Human evaluation} presents the human evaluation results on a randomly sampled test set. 
    200 samples were evaluated and the order of evaluation systems were shuffled. 
    Three graduate students were asked to score from 0 to 3 (3 for the best) and the average scores were reported. 
    The Fleiss’s kappa value~\cite{fleiss1971measuring} for each model was also reported, indicating the inter-judge moderate agreement during evaluation.
    It can be seen that HeterMPC$_{\emph{BERT}}$ and HeterMPC$_{\emph{BART}}$ achieved higher subjective quality scores than the baselines. Their kappa values were also higher than the BERT and BART baselines, respectively.

  \subsection{Analysis}

    \begin{figure}[t]
      \centering
      \includegraphics[width=6.5cm]{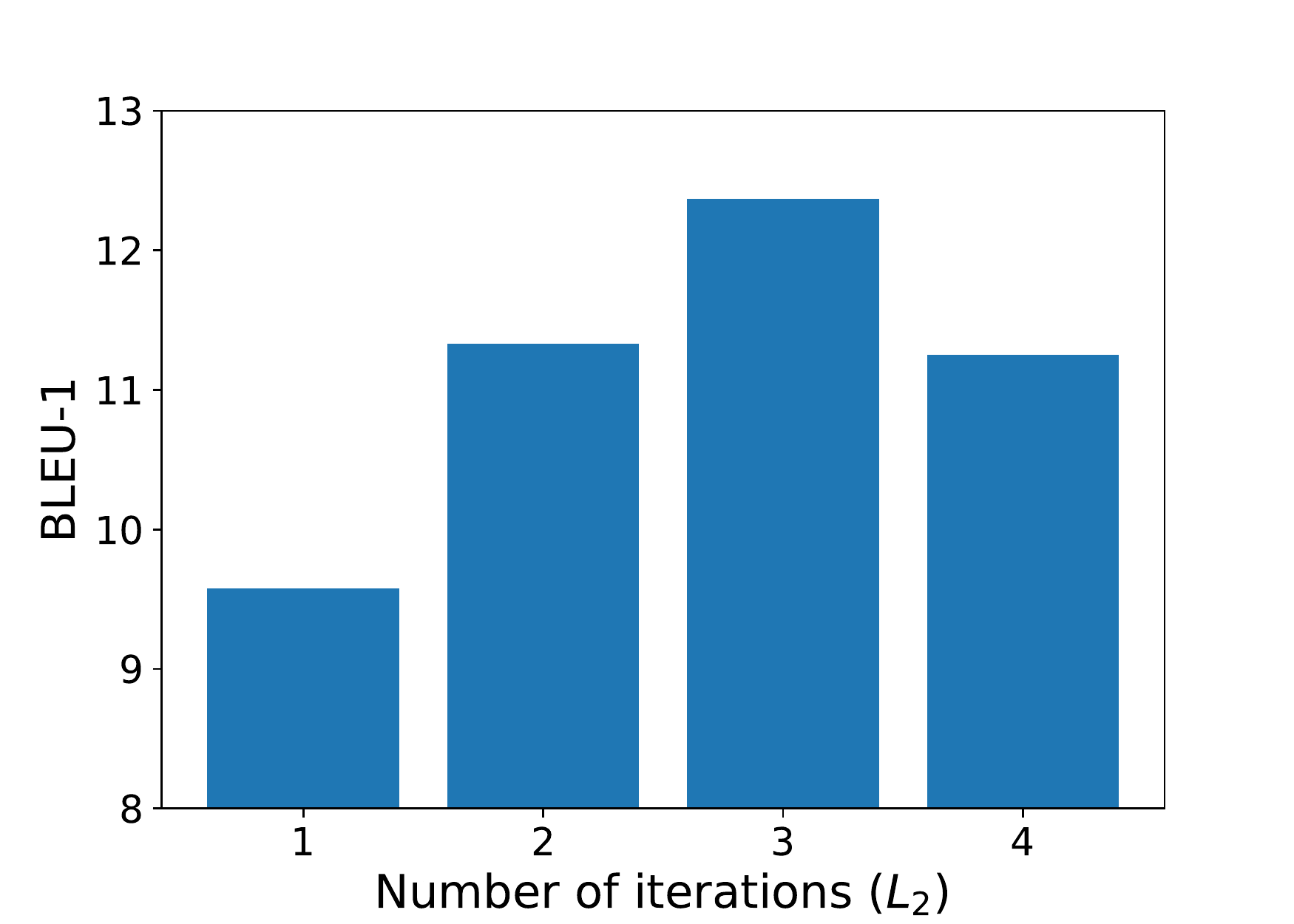}
      \caption{Performance of HeterMPC$_{\emph{BERT}}$ under different numbers of iterations ($L_2$) on the test set.}
      \label{fig-layers-l2}
    \end{figure}
  
    \paragraph{The impact of numbers of iterations ($L_2$).}
    Figure~\ref{fig-layers-l2} illustrates how the performance of HeterMPC$_{\emph{BERT}}$ changed with respect to different numbers of iterations ($L_2$) on the test set. 
    It can be seen that the performance of HeterMPC$_{\emph{BERT}}$ was significantly improved as $L_2$ increased at the beginning, which shows the effectiveness of incorporating the contextual information between nodes. 
    Then, the performance was stable and dropped slightly. 
    The reason might be that models begin to overfit due to a larger set of parameters.

    \begin{figure}[t]
      \centering
      \includegraphics[width=6.5cm]{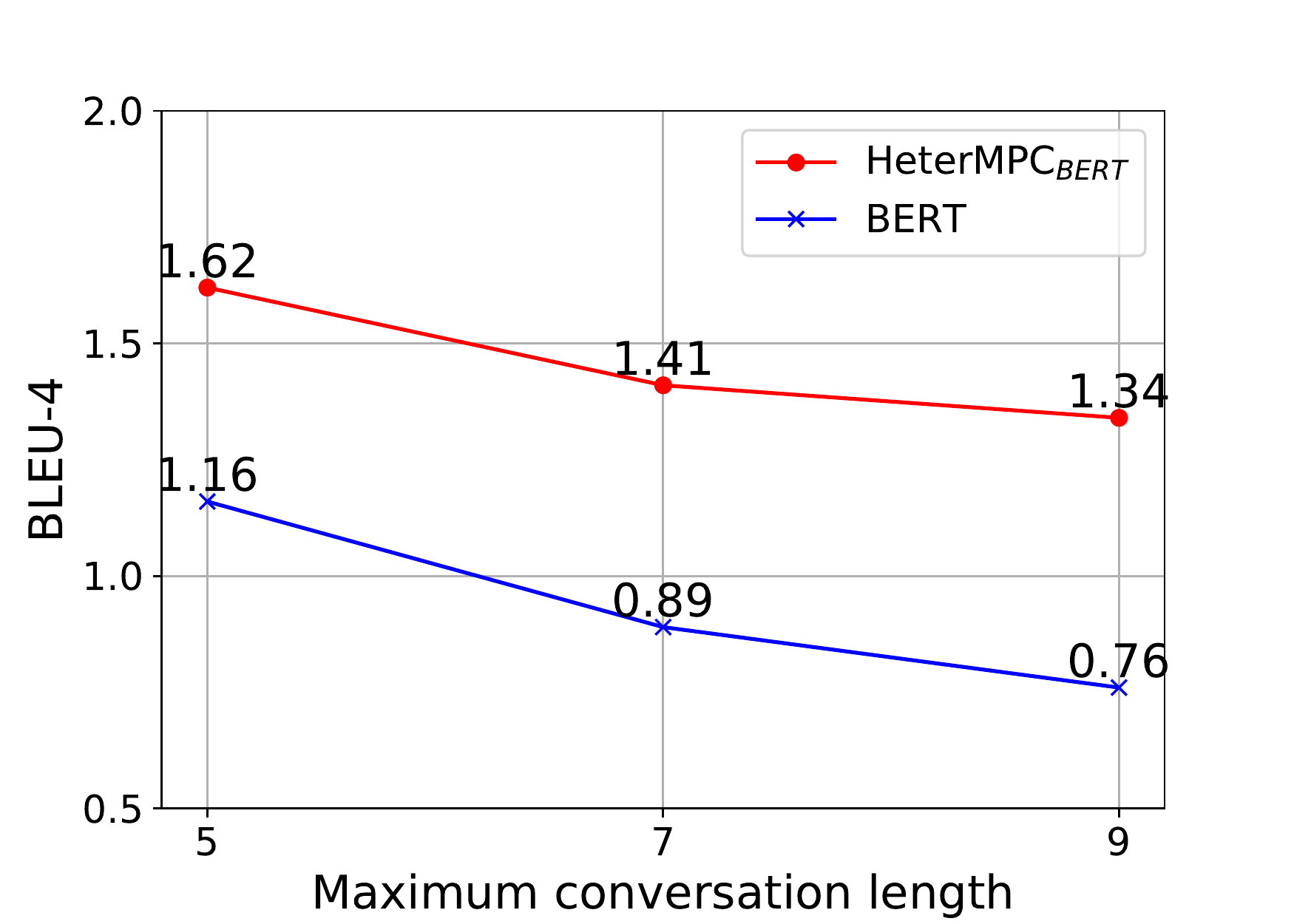}
      \caption{Performance of BERT and HeterMPC$_{\emph{BERT}}$ on test samples with different session lengths.}
      \label{fig-conv-length}
    \end{figure}

    \paragraph{The impact of conversation length.~\footnote{This experiment can also be considered as exploring the impact of interlocutor number, since more context utterances indicate that more interlocutors are involved implicitly.}}
    Figure~\ref{fig-conv-length} illustrates how the performance of HeterMPC$_{\emph{BERT}}$ changed according to the test samples with different session lengths.
    As the session length increased, the performance of HeterMPC$_{\emph{BERT}}$ dropped less than that of BERT,
    showing superiority of our method on dealing with longer conversations.
    
  \begin{table}[!hbt]
    \small
    \centering
    \setlength{\tabcolsep}{4.0pt}
    \begin{tabular}{c|p{4.5cm}|c}
    \toprule
      \textit{Speaker}      &  \multicolumn{1}{c|}{\textit{Utterance}}              &  \textit{Addressee}            \\ 
    \midrule
                              \multicolumn{3}{c}{Case 1}   \\ 
    \hline
      I.1                   &  geev: in gparted now?                                &    -                  \\ 
    \hline
      \multirow{2}{*}{I.2}  &  there is no such command in my computer              &  \multirow{2}{*}{I.1}   \\ 
    \hline
      \multirow{2}{*}{I.1}  &  open a terminal, type: sudo apt-get install gparted  &  \multirow{2}{*}{I.2}  \\ 
    \hline
      \multirow{2}{*}{I.1}  &  after: gparted aksks for your pw, opens a gui        &  \multirow{2}{*}{I.2}  \\ 
    \hline
      I.2                   &  ok if found fail i will contact you                  &  I.1  \\ 
    \hline
     \multirow{12}{*}{I.3}  &  does it say ``to install type sudo  apt-get install gparted''? \textbf{(Human)} &  \multirow{12}{*}{I.2}  \\ 
    \cline{2-2}
                            &  do you have remote access to remote desktop? \textbf{(GSN)}   &  \\
    \cline{2-2}
                            &  ok, i will try that. \textbf{(BERT)}                 &    \\ 
    \cline{2-2}
                            &  you can use the command ``sudoapt-get install gparted'' to install the gparted livecd. \textbf{(HeterMPC$_{\emph{BERT}}$)} &  \\
    \cline{2-2}
                            &  i'm not sure what you mean by ``in gparted''  \textbf{(BART)} &  \\
    \cline{2-2}
                            &  you need to install gparted, it is a gui tool \textbf{(HeterMPC$_{\emph{BART}}$)} &  \\
    \hline
                              \multicolumn{3}{c}{Case 2}   \\ 
    \hline
      \multirow{4}{*}{I.1}  &  got a bit of a mount problem. fstab is still the same but the  secondary harddrive shows as  busy when doing mount  & \multirow{4}{*}{-}   \\ 
    \hline
      \multirow{2}{*}{I.2}  &  you can see what is mounted with : mount                &  \multirow{2}{*}{I.1} \\
    \hline
      I.1                   &  yes it is not mounted                                   &  I.2  \\
    \hline
      \multirow{2}{*}{I.3}  &  how would i check what module the cam uses?             &  \multirow{2}{*}{I.2}  \\
    \hline
      \multirow{2}{*}{I.2}  &  lsmod less will show you te loaded modules              &  \multirow{2}{*}{I.3}  \\
    \hline
      \multirow{2}{*}{I.1}  &  lmount - t ext3 filepath fails  cause already mounted or busy               &  \multirow{2}{*}{I.2}  \\
    \hline
      \multirow{11}{*}{I.2}  &  then try to mount it at command line \textbf{(Human)}   &  \multirow{11}{*}{I.1}  \\
    \cline{2-2}
                            &  i'm not sure how to do that \textbf{(GSN)}              &                        \\
    \cline{2-2}
                            &  i'm not sure what the problem is \textbf{(BERT)}        &                        \\
    \cline{2-2}
                            &  you need to mount it as a mount  point \textbf{(HeterMPC$_{\emph{BERT}}$)}&       \\
    \cline{2-2}
                            &  i'm not sure what the problem is \textbf{(BART)} &  \\
    \cline{2-2}
                            &  you need to check the filepath file \textbf{(HeterMPC$_{\emph{BART}}$)} &  \\
    \bottomrule
    \end{tabular}
    \caption{The response generation results of two test samples.
    ``I." is the abbreviation of ``interlocutor".
    We kept original texts without manual corrections.
    }
    \label{tab: Generation}
  \end{table} 

    \paragraph{Case Study.}
    Case studies were conducted by randomly sampling two MPC instances as shown in Table~\ref{tab: Generation}. 
    Given the conversation graph of the first case, the response to generate addresses I.2. 
    Thus, the information relevant to I.2 should be collected.
    We can see that ``\emph{gparted}'' in the first utterance is two hops away from I.2 (the first utterance is replied by the second utterance which is spoken by I.2), and this word in the fourth utterance and ``\emph{install gparted}'' in the third utterance are both one hop away from I.2 (these two utterances directly address I.2). 
    The responses generated by HeterMPC$_{\emph{BERT}}$ and HeterMPC$_{\emph{BART}}$ both contain these keywords, showing that it can capture the conversation graph information accurately and generate a human-like response. 
    However, due to the lack of the interlocutor information and the conversation structure, GSN generated an irrelevant response. 
    BERT generated a response which seems replying to the third utterance. 
    Although BART captured ``\emph{gparted}'', it failed to handle the action ``\emph{install}''.
    In the second case, we can see that the responses generated by GSN, BERT and BART are general and useless while HeterMPC$_{\emph{BERT}}$ and HeterMPC$_{\emph{BART}}$ can still generate a suitable response. 
    Due to the complicated interactions between utterances and interlocutors, the conversation flow might be led by some unnecessary information, which shows the importance of making models aware of the conversation structure.

    
    
  
    \paragraph{Robustness.}
    The addressee labels are important for constructing a graph used in HeterMPC.
    This kind of label is commonly available in real life such as “A@B” labels in group chatting, Twitter, Reddit and various forums that denote speaker A talking to addressee B. 
    However, addressee labels of a part of utterances are missing in the existing MPC datasets since a speaker may forget to specify an addressee.
    HeterMPC is robust since utterances without addressee labels can be assigned with a general addressee label “\emph{To all interlocutors}”. 
    We leave evaluation on other datasets in future work.
    

\section{Conclusion}
  We present HeterMPC to model complicated interactions between utterances and interlocutors in MPCs with a heterogeneous graph. 
  Two types of graph nodes and six types of edges are designed.
  Node-edge-type-dependent parameters are introduced for better utilizing the structural knowledge of conversations during node updating.
  Results show that HeterMPC outperforms baselines by significant margins, achieving a new state-of-the-art performance for response generation in MPCs on the Ubuntu IRC benchmark. 
  In the future, we will explore better ways of maximizing feature distribution differences to model heterogeneity.

\section*{Acknowledgements}
  We thank anonymous reviewers for their valuable comments.

\clearpage
\bibliography{custom}
\bibliographystyle{acl_natbib}


\end{document}